\PassOptionsToPackage{unicode}{hyperref}
\PassOptionsToPackage{hyphens}{url}
\PassOptionsToPackage{dvipsnames,svgnames,x11names}{xcolor}
\documentclass[
  11pt,
]{article}
\usepackage{xcolor}
\usepackage[margin=1in]{geometry}
\usepackage{amsmath,amssymb}
\usepackage{iftex}
\ifPDFTeX
  \usepackage[T1]{fontenc}
  \usepackage[utf8]{inputenc}
  \usepackage{textcomp} 
\else 
  \usepackage{unicode-math} 
  \defaultfontfeatures{Scale=MatchLowercase}
  \defaultfontfeatures[\rmfamily]{Ligatures=TeX,Scale=1}
\fi
\usepackage{lmodern}
\ifPDFTeX\else
\fi
\IfFileExists{upquote.sty}{\usepackage{upquote}}{}
\IfFileExists{microtype.sty}{
  \usepackage[]{microtype}
  \UseMicrotypeSet[protrusion]{basicmath} 
}{}
\makeatletter
\@ifundefined{KOMAClassName}{
  \IfFileExists{parskip.sty}{%
    \usepackage{parskip}
  }{
    \setlength{\parindent}{0pt}
    \setlength{\parskip}{6pt plus 2pt minus 1pt}}
}{
  \KOMAoptions{parskip=half}}
\makeatother
\usepackage{longtable,booktabs,array}
\usepackage{calc} 
\usepackage{etoolbox}
\makeatletter
\patchcmd\longtable{\par}{\if@noskipsec\mbox{}\fi\par}{}{}
\makeatother
\IfFileExists{footnotehyper.sty}{\usepackage{footnotehyper}}{\usepackage{footnote}}
\makesavenoteenv{longtable}
\usepackage{graphicx}
\makeatletter
\newsavebox\pandoc@box
\newcommand*\pandocbounded[1]{
  \sbox\pandoc@box{#1}%
  \Gscale@div\@tempa{\textheight}{\dimexpr\ht\pandoc@box+\dp\pandoc@box\relax}%
  \Gscale@div\@tempb{\linewidth}{\wd\pandoc@box}%
  \ifdim\@tempb\p@<\@tempa\p@\let\@tempa\@tempb\fi
  \ifdim\@tempa\p@<\p@\scalebox{\@tempa}{\usebox\pandoc@box}%
  \else\usebox{\pandoc@box}%
  \fi%
}
\def\fps@figure{htbp}
\makeatother
\usepackage{framed}
\usepackage{pdflscape}
\usepackage{microtype}
\usepackage{bookmark}
\IfFileExists{xurl.sty}{\usepackage{xurl}}{} 
\hypersetup{
  pdftitle={The novelty exposed: ablation{,} reproducibility and transparency in 40 technical reports of open-weight large language models (LLMs)},
  pdfauthor={Bahaeddin Türkoğlu},
  colorlinks=true,
  linkcolor={blue},
  filecolor={Maroon},
  citecolor={Blue},
  urlcolor={blue},
  pdfcreator={LaTeX via pandoc}}

\title{The novelty exposed: ablation, reproducibility and transparency
in 40 technical reports of open-weight large language models (LLMs)}
\author{Bahaeddin Türkoğlu\thanks{Department of Artificial Intelligence and Data Engineering, Ankara University, Ankara, Türkiye. ORCID: \url{https://orcid.org/0000-0003-0255-8422}. Correspondence: turkoglub@ankara.edu.tr}}
\date{}

\begin{document}
\maketitle
\begin{abstract}
In 2025 alone, 48 new large language model (LLM) architectures were
added to the Hugging Face transformers library, and many of them arrive
with a technical report that presents the model as new. What is actually
new in these models, and is it tested? We put this question to LLM
technical reports. We look at three things: what the reference code
shows about how new open-weight architectures differ from earlier ones;
what 40 randomly sampled technical reports claim as new and whether they
support it with ablation studies; and what they disclose for
reproducibility and transparency. In the library, 39 of the 48
architectures added in 2025 take at least two of their three core blocks
from another model, most often Llama, although part of this reflects how
the library writes code. In the reports, 19 of 73 architectural or
training novelty claims come with a controlled with/without comparison;
4 of 40 reports state more than one training run; 8 rerun the models
they compare against under one setup; and 1 discloses data mixture,
token count, and compute together. We also looked for known components
given new names: none of five candidate cases hid its origin, though we
did not study model names. We propose seven reporting rules for LLM
release reports and apply them to two reports. We do not judge model
quality, and an untested claim may still be true.

\textbf{Keywords:} large language models; LLM technical reports;
ablation study; reproducibility; transparency; LLM evaluation;
open-weight models
\end{abstract}

\subsection{1. Introduction}\label{introduction}

This paper asks a plain question of the release reports of large
language models. What does a report claim as new, and what evidence does
it give that the new part helps? The question is not new. In research on
metaheuristics, Sörensen asked whether methods presented as new were in
fact new (Sörensen, 2015), and studies of particular algorithms answered
it by examining what those algorithms do (Weyland, 2012;
Camacho-Villalón et al., 2023). Those studies asked what a method does,
what is new about it, and whether the evidence supports that claim.

We put these questions to language-model releases because of a problem
of attribution, which is knowing which change produced a model's
performance. Narang and colleagues tested many proposed Transformer
modifications in one shared implementation and found that most did not
improve performance there (Narang et al., 2021). A follow-up reports the
same at 1 to 3 billion parameters (Zhao et al., 2026). A change that
helps in the paper that proposes it may therefore not help elsewhere. A
claim about a particular change needs a comparison that tests that
change.

A release report does several jobs at once. It describes a system,
reports results, explains how the system was trained and names the
choices its authors consider new. The model comes from many decisions:
the architecture, the training data, the amount of training and the
training procedure. A comparison between complete systems tells a reader
which one performed better under the stated conditions. When several
conditions change together, it says little about any one of them. We do
not assume that novelty must sit in one part of a system rather than
another. We ask whether a report identifies its claimed contribution and
supplies evidence that bears on it.

None of this denies the progress of the field. The transformer, scaling
laws, instruction tuning with human feedback, mixture of experts and
test-time compute are real advances, each set out and studied as a
method (Vaswani et al., 2017; Kaplan et al., 2020; Ouyang et al., 2022;
Shazeer et al., 2017; Snell et al., 2024). Releases that build on them,
such as Llama, made capable models widely available (Touvron et al.,
2023). A release can be a useful contribution when it documents a model
and its performance, even if it is not an experiment built to isolate a
new idea. But when a report presents a design choice as new, a reader
should be able to tell that claim apart from the result of the complete
system.

Our main results are short to state. We coded 40 release reports, drawn
as a random sample, stratified by year, from the 64 eligible reports
that the documentation of the Hugging Face transformers library links
to. Twelve of the 40 make no architectural or training novelty claim,
and the other 28 make 73 such claims. Of these 73, 19 (26\%) come with a
controlled with/without comparison at matched size, data, and tokens.
Another 20 (27\%) come with a comparison that changes other things at
the same time, and 34 (47\%) come with no comparison. Counted by report,
11 of the 28 test at least one claim in a controlled way, and 4 test all
of them. We see no clear trend over time. We also scored each report on
five items our coding measures, and no report meets more than three of
them, while 24 of the 40 meet none. These counts describe the evidence
in the reports, not the quality of the models, and an untested claim may
still be true.

We also looked for places where a report gives a new name to a technique
that it says is already known. The two coders proposed five candidate
cases, and on reading the quoted passages none hid the origin of the
component. We did not study model names. The gap we did find lies
elsewhere, in the distance between what a report presents as new and
what it tests.

Section 2 traces how the field arrived at its current models, and
Section 3 examines first the library code and then what the reports
claim and test. Section 4 asks why weak evidence persists, Section 5
proposes rules for release reports, and Section 6 states our conclusions
and the limits of the study.

\subsection{2. A short history of new language
models}\label{a-short-history-of-new-language-models}

The field has real advances to its name, and most of them arrived as
methods in research papers. In 2017 the transformer replaced recurrence
with attention (Vaswani et al., 2017). Attention lets a model draw on
information from other positions in a sequence when it processes a word.
The paper compared the new architecture with earlier translation models.
It also varied its own components, such as the number of attention
heads, to see what each contributed.

BERT, reported in 2019, pre-trained a transformer using context from
both directions and then adapted it to individual tasks (Devlin et al.,
2019). Its authors removed parts of the training objective one at a time
and reported how performance changed. The GPT-3 work in 2020 examined
what a language model could do from examples supplied in its input,
without a separate training step for each task (Brown et al., 2020). It
trained the same design at a range of sizes, so that the effect of size
could be read from the results.

A second line of work asked how to train larger models. In 2020, work on
scaling laws described how performance varies with model size, data and
computation (Kaplan et al., 2020). The relationships were fitted to many
training runs that varied these quantities. In 2022, work on
compute-optimal training asked how a fixed budget should be divided
between model size and data (Hoffmann et al., 2022). Its authors trained
models of many sizes on different amounts of data. They then tested
their conclusion by training a smaller model on more data with the same
compute as a larger one, and comparing the two.

In the same year, instruction tuning with human feedback addressed how
to make models follow what users ask of them (Ouyang et al., 2022). The
method fine-tunes a model on demonstrations and then on human judgements
of its outputs. The test was direct: human raters compared outputs of
the tuned models with those of the models they started from. These three
papers concern different choices: how much to train, on what data, and
toward what behaviour.

Mixture of experts changed model design. A selection mechanism activates
only part of the network for each input, so a model can hold more
parameters than it uses at any one time (Shazeer et al., 2017). The 2017
paper compared such layers with dense models on language modelling and
translation. Switch Transformers, published in 2022, simplified the
selection to a single expert per input and developed the method for
transformer models (Fedus et al., 2022). Its comparisons were made
against dense models that spend the same computation on each token.

Work published in 2024 turned to the computation a model spends when it
answers a question, rather than during training (Snell et al., 2024). It
compared ways of spending that computation and found that the best
allocation can depend on how hard the problem is. It also set extra
computation at answer time against a larger model, and reported when one
was the better use of resources than the other.

The Llama release in 2023 belongs to a different kind of work. It
presented a family of open and efficient foundation language models,
trained on publicly available data (Touvron et al., 2023). Its
architecture combined choices that earlier papers had proposed, and its
report says so. Llama then became a widely used base for later open
models. Section 3.1 shows how often its code is reused. Its value lay in
making a capable model available to people outside the company that
trained it. That is a contribution of its own, but it is not the same as
introducing a new method.

The two kinds of work answer to different evidence. Each paper above
stated a method and then ran an experiment designed to show what that
method did, against a baseline that differed in that respect. A release
shows what a complete model can do. When a release also presents one of
its parts as new, the question is whether it supplies evidence of the
first kind for that part.

\subsection{3. What the code shows and what the reports
claim}\label{what-the-code-shows-and-what-the-reports-claim}

\subsubsection{3.1 The code}\label{the-code}

\begin{figure}
\centering
\includegraphics[width=1\linewidth,height=\textheight,keepaspectratio,alt={Figure 1. Causal language-model architectures added to the Hugging Face transformers library each year (grey), those whose code takes at least two of the three core blocks (attention, feed-forward, decoder layer) from another model (blue, with count and share), and those taking at least one core block from Llama (red line). 2026 runs to 13 November. The dashed line marks the library's move, from about 2024, to writing new models as edits of older ones; part of the later rise reflects that change.}]{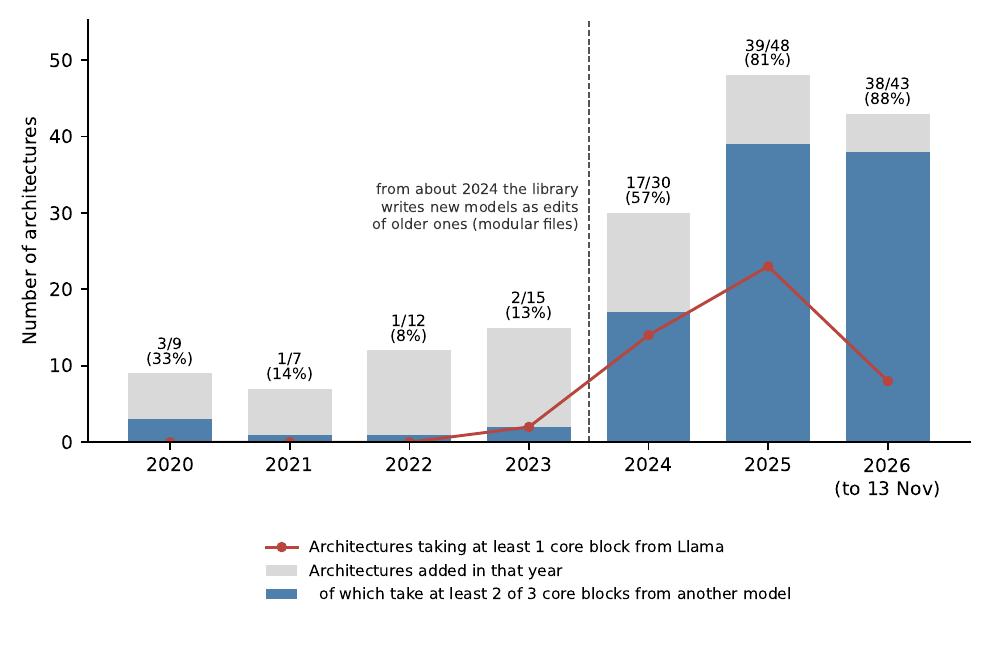}
\caption{Figure 1. Causal language-model architectures added to the
Hugging Face transformers library each year (grey), those whose code
takes at least two of the three core blocks (attention, feed-forward,
decoder layer) from another model (blue, with count and share), and
those taking at least one core block from Llama (red line). 2026 runs to
13 November. The dashed line marks the library's move, from about 2024,
to writing new models as edits of older ones; part of the later rise
reflects that change.}
\end{figure}

\begin{figure}
\centering
\includegraphics[width=1\linewidth,height=\textheight,keepaspectratio,alt={Figure 2. The attention class of Qwen3 against Llama in the reference code (HF transformers, commit 0d7a621). Shaded statements differ.}]{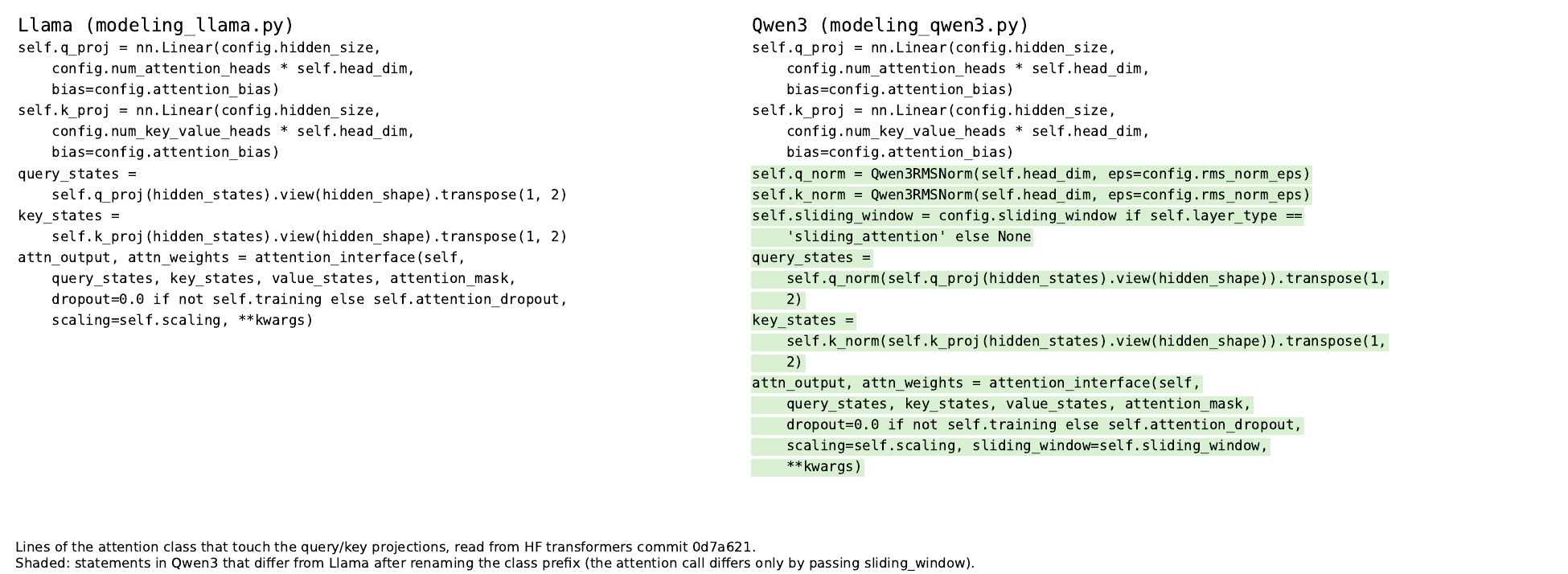}
\caption{Figure 2. The attention class of Qwen3 against Llama in the
reference code (HF transformers, commit 0d7a621). Shaded statements
differ.}
\end{figure}

A release report is one record of a model. The reference code in the
Hugging Face transformers library is another (Wolf et al., 2020). It
does not tell us how a model was designed or trained. It does show, in a
form anyone can read, which parts of one implementation are written as
parts of another. We start there because it is the most direct evidence
of what one model shares with another.

Our population is the 164 causal language-model architectures in the
library that have a date of addition. For each year, we count the
architectures whose code takes at least two of three core blocks from
another model. The three blocks are attention, feed-forward and the
decoder layer. The attention and feed-forward blocks do the model's main
computation, and the decoder layer combines them into one step of
processing. The unit is a dated library architecture, not a paper or a
vendor announcement.

The early years have small counts. In 2020, 3 of 9 architectures met the
criterion (33\%); in 2021, 1 of 7 (14\%). In 2022, the figure was 1 of
12 (8\%), and in 2023, 2 of 15 (13\%). With counts this small, the
classification of one architecture can move a yearly share a long way.

From 2024, the share rises. In 2024, 17 of 30 architectures met the
criterion (57\%); in 2025, 39 of 48 (81\%). The figure for 2026 covers
only the part of the year up to 13 November: 38 of 43 (88\%). Figure 1
gives the yearly shares.

Part of this rise comes from how the library writes code. Around 2024,
the library began to write new models as edits of older ones, in what it
calls modular files. Under that practice, taking a block from an
existing model is the expected way to add a new one. Part of the rise
after 2023 therefore reflects how the code is written, not only how
models are designed. Our data cannot divide the rise between the two,
and we do not try.

The same code can be read from the lender's side. Across the library,
288 core blocks are taken from another model. Llama lends the most: 80
blocks, used by 47 architectures (Touvron et al., 2023). The next
lenders are DeepSeek-V3 with 13 blocks, Qwen3-MoE with 12 and Mistral
with 10. The yearly shares describe the architectures that borrow, and
these counts describe where the borrowed blocks come from. Both measure
code reuse, not design influence.

To see what such reuse looks like at the level of single lines, we
compared five models with Llama in the reference code. Figure 2 shows
one of the comparisons. The comparison shows where the reference code of
each model differs from Llama's. We chose the five as illustrations,
because they are well-known open models whose library code is written as
a Llama variant. They are not a sample, and they say nothing about how
often architectures differ by a given amount.

In this code, Qwen3 differs from Llama by QK-norm, which the Qwen3
report describes as a normalisation in the attention mechanism (Yang et
al., 2025). One normalisation is applied to the queries and one to the
keys. In the library code the change takes 2 added lines and 2 changed
lines, plus an optional sliding-window switch that is off by default
(Figure 2) (Wolf et al., 2020). Qwen2 differs from Llama by turning on
bias in the query, key and value projections (Yang et al., 2024).
Mistral differs by a sliding-window attention mask, which limits the
positions each token can attend to (Jiang et al., 2023). OLMo 2 moves
the normalisation from the input to the output of each sub-layer and
adds QK-norm (OLMo et al., 2025). Gemma uses a GeGLU gate, scales the
embeddings and parametrises the RMSNorm weight differently (Gemma Team
et al., 2024). These are differences in the reference code. They are not
every choice that went into the released models, which also differ in
data, training, and scale.

Several of these parts were published before the models that use them.
Sparse or sliding-window attention appeared in 2019 and 2020,
multi-query attention in 2019, and GLU variants in 2020 (Child et al.,
2019; Beltagy et al., 2020; Shazeer, 2019; Shazeer, 2020). Grouped-query
attention and QK-norm followed in 2023 (Ainslie et al., 2023; Dehghani
et al., 2023). An earlier source does not settle what a part contributes
to a later model. It does let a reader separate where a part came from
and what it is said to do in the new system.

A short code difference is not an equivalence result. It shows where a
model differs from another, not that the model is not new. Two changed
lines can matter a great deal in training, and the code cannot tell us
whether they do. What the comparison does is locate the change. Once a
change is located, it is clear which claim needs evidence and what that
evidence would have to isolate.

In the reports whose attribution we read, the vendors credit these
parts. The Qwen3 report states: ``we remove QKV-bias used in Qwen2
\ldots{} and introduce QK-Norm (Dehghani et al., 2023) to the attention
mechanism to ensure stable training'' (Yang et al., 2025; Dehghani et
al., 2023). The Qwen2 report states: ``We adopt Grouped Query Attention
(GQA, Ainslie et al., 2023) instead of conventional multi-head
attention'' (Yang et al., 2024; Ainslie et al., 2023). A citation of
this kind tells the reader where a part came from. It does not tell the
reader what the part adds to the new model. The OLMo 2 report does both:
it cites the origin of each architectural change and shows its effect in
a figure (OLMo et al., 2025).

\subsubsection{3.2 Names}\label{names}

A known component under a new name makes it harder for a reader to see
where that component came from. We therefore looked for places where a
release report gives a new name to a technique that the report itself
says is already known.

Across 40 reports, the two coders proposed five cases. We read the
quoted passages, and none hides the origin of the technique. Falcon
calls grouped-query attention ``multigroup'', but in the same sentence
it gives the name ``grouped query attention'' and credits its source
(Almazrouei et al., 2023; Ainslie et al., 2023). Falcon Mamba says that
``the standard method for prefill'' is ``referred to here as Parallel
Prefill'' (Zuo et al., 2024). Mamba says it uses ``selection'' instead
of ``gating'' because the older term is overloaded (Gu and Dao, 2023).
Each of these states what it renames, and Mamba also says why.

The other two candidates are not renamings. Mamba-2 defines a
state-space version of multi-query attention, which is a new analogue
and not a new name for an old technique (Dao and Gu, 2024). Falcon's
``monolayer'' is linked to a known technique only by the coders'
reading, not by anything in the report (Almazrouei et al., 2023).

Among these five candidates, then, we found no hidden renaming of a
known component. The search covered only the passages the coders
proposed, so other cases may exist in the same reports. It also covered
only components. We did not study model names. Whether a new model name
presents a known design as something else is a question this study
leaves open.

\subsubsection{3.3 Claims and tests}\label{claims-and-tests}

We coded 40 release reports. They are a random sample, stratified by the
year each report appeared on arXiv, from the 64 eligible reports that
the library documentation links to. A linked paper was eligible if it
named the model, dated from 2020 or later, and was not a paper about one
component. Appendix A gives the method in full; we describe it briefly
here.

Two language models from different families, Claude Opus 5.5 and OpenAI
GPT-6.1 Sol, coded every report independently with the same
instructions. For every value they gave a verbatim quote and a page. A
script checked every quote against the PDF text. If the quote for a
coded value was not found, the value fell to the lowest level; if the
quote for a claim was not found, the claim was dropped. The script
confirms that a quote exists in the report. It does not confirm that the
coder read the quote correctly.

The two coders listed claims separately, so their lists had to be
matched before they could be compared. We matched claims by wording and
page. This gave 51 matched claims, whose ablation levels we compared
directly. Another 53 claims were listed by only one coder. These went to
a third model family, xAI Grok 4.7, together with every disagreement on
matched claims, and its quotes were checked the same way. Agreement on
the ablation level of matched claims, measured by Cohen's kappa, was
0.67 (Table A1). On runs, rerunning, copied scores, third-party
evaluation, tokens and compute it ranged from 0.73 to 0.88. Because the
unmatched claims were settled by the third coder, part of the final
claim count and its ablation levels rests on one model's reading. Of the
73 final claims, 40 were coded the same way by both coders and 33 were
settled by the third model; of the 19 controlled claims, 12 come from
agreement and 7 from the third model.

We counted a contribution as a novelty claim when the authors stated it
as new in the abstract, the list of contributions, or the introduction.
The definition takes each report at its word. We do not decide for
ourselves whether an idea deserves to be called new, and we do not treat
every implementation detail as a claim.

Twelve of the 40 reports make no architectural or training novelty
claim. The other 28 make 73 such claims. Of these, 19 (26\%) come with a
controlled with/without comparison at matched model size, data, and
tokens. Another 20 (27\%) come with a comparison that changes other
things at the same time, and 34 (47\%) come with no comparison.

Counted by report, the picture is somewhat different. Of the 28 reports
that make such claims, 11 test at least one claim in a controlled way,
and 4 test all of them. So a controlled test is not rare at the level of
reports. A report that tests every claim it makes is.

Claims cluster within reports, which is why the two counts differ. In
2025, one report, LongCat-Flash, makes 9 of the 27 claims, and none of
the 9 is controlled. A single report can therefore move a yearly share a
long way.

By year of the report, 3 of 10 claims were controlled in 2020-2022, 2 of
5 in 2023, 11 of 28 in 2024, 3 of 27 (11\%) in 2025 and 0 of 3 in 2026.
By report, at least one claim was controlled in 2 of 6 reports in
2020-2022, 1 of 3 in 2023, 5 of 10 in 2024, 3 of 8 in 2025 and 0 of 1 in
2026. The low claim-level share for 2025 owes much to one report, and
the report-level figures for that year are close to those of earlier
years. We see no clear trend over time.

The three categories of evidence answer different questions. A matched
comparison changes the claimed part and holds size, data, and tokens
fixed. A difference in the result can then be read against that part, in
that setting. It need not hold in another implementation or application,
and earlier work found that many proposed Transformer modifications did
not carry over (Narang et al., 2021; Zhao et al., 2026). Even so, it is
the only one of the three that tests the claim as stated.

A comparison that changes several things at once still answers a
practical question: which complete model did better under the reported
conditions. It cannot say which change made the difference. That matters
when a report credits a gain to one architectural or training choice.
The complete model is a result of the work, but a claim about a part
needs evidence about that part. A claim with no comparison may still
describe a method well and come with code that others can test. Within
the report, though, it stands untested, and an untested claim may still
be true.

Repeated runs address a different uncertainty. Results vary across
training runs that differ only in chance events such as the random seed,
so a spread across runs tells the reader how much weight a difference
can bear (Dodge et al., 2019; Bouthillier et al., 2021). Four of the 40
reports state more than one training run or report a spread across runs.
None of the 26 reports from 2024 onward does.

The cost is real. At full pre-training scale, each extra run costs a
full training budget, and we do not expect every team to pay it. The
minimum is cheaper. A report can say that a result comes from a single
run, so that the reader knows no spread was measured.

\subsubsection{3.4 The yardstick}\label{the-yardstick}

A new model is usually shown against other models. Which models are
chosen matters, and so does how their scores were obtained. A common
evaluation setup measures every model with the same yardstick. Scores
copied from separate reports may have been produced under different
conditions, and the comparison is then harder to read (Melis et al.,
2018).

Eight of the 40 reports reran the models they compared against under one
evaluation setup. The Mistral 7B report is an example: it reran all
comparison models with its own evaluation pipeline (Jiang et al., 2023).
This does not settle every question about evaluation, but the reader
knows where each score came from.

Twenty-two of the 40 reports took some or all comparison scores from
other reports. Sometimes that is the only option, because another model
cannot be run. The reader then needs to know which scores were copied.
Otherwise, a difference may belong to the models, to the evaluation
procedures, or to both, and the report gives no way to tell.

Eight of the 40 reports include any evaluation run by a third party.
Such an evaluation is evidence that the releasing team did not produce
itself. Like any score, it still needs enough description for the reader
to know what was measured.

\subsubsection{3.5 Disclosure}\label{disclosure}

The data mixture, the token count and the compute each answer a
different question about how a model was made. One does not stand in for
another.

Two of the 40 reports describe the training-data mixture fully enough to
rebuild it. The test was whether another party could assemble the
mixture from the description, not whether the report named broad
sources. Coder agreement on this field was low (kappa 0.30), so the
count of 2 is the least certain number in the paper.

Twenty-nine of the 40 reports give the number of pre-training tokens,
and 12 of the 40 give the compute used. A token count says how much data
was processed. It does not say what the data were or how the sources
were chosen. Compute says what resources stand behind a result and what
it would take to reproduce or match it. These counts record what the
reports state, not what their authors knew or were free to disclose.

We read attribution in seven reports: Qwen2, Qwen3, OLMo 2, Falcon,
Falcon Mamba, Mamba and Mamba-2. Each credits the sources of the parts
we looked at. In these reports, the weaker link is elsewhere. It lies
between a claim and a test of it, and between a comparison and the
conditions under which its scores were produced.

\subsection{4. Why weak evidence
persists}\label{why-weak-evidence-persists}

A model release bundles several kinds of work. A team may adopt an
existing architecture, change part of it, assemble training data, choose
a training procedure and evaluate the result. Each choice can matter to
the final model. The trouble starts when a report presents the bundle as
a contribution without saying which parts are new and what each one did.
A strong result for the bundle says little about any single change
inside it. This is a problem of evaluation. Building on earlier work is
not the problem.

Two questions are easy to confuse here. A comparison between complete
models tells a reader what a release can do. A controlled comparison
asks whether a stated change helped when the other conditions were held
fixed. A user deciding whether to adopt a model needs the first answer.
A researcher deciding which idea to build on needs the second. A report
can serve the user well and give the researcher little to go on. Some
reports keep the two apart. The OLMo 2 report cites the origin of each
architectural change and shows its effect in a figure, and the Mistral
7B report reran all comparison models with its own evaluation pipeline.
Whether a report does this is left to its authors.

The usual outside check on a claim is peer review. All 40 reports in our
sample were taken from arXiv as preprints. Release reports as such are
not peer reviewed before they appear. We did not check whether some of
the 40 later appeared at a reviewed venue, so we cannot say how many
were reviewed after release. Teams may also ask colleagues for comments
before a release, and we have no record of that. Review does not
guarantee a sound comparison, and an unreviewed report can be careful
work. But when a report is first read, the question of whether its
evidence bears on its claim has usually not been put to anyone outside
the team. Readers are left to ask it.

Section 3 showed what readers then find. We counted the novelty claims
that come with a controlled comparison, those that come with a
comparison changing several things at once, and those that come with
none. The second and third kinds are not failures of honesty. They are
limits on what the report can show. Where a comparison changes several
things at once, its result cannot be assigned to any one of them. Where
there is no comparison, the claimed effect is untested in the report,
and an untested claim may still be true.

Untested claims would matter less if proposed changes usually worked.
Narang and colleagues tested many proposed Transformer modifications in
one shared implementation and found that most did not improve
performance there, and a follow-up reports the same at 1 to 3 billion
parameters (Narang et al., 2021; Zhao et al., 2026). Those results do
not make any claim in our sample false. They do mean that the presence
of a change in a successful model is weak evidence that the change
helped.

The library itself makes reuse both easy and visible. As Section 3.1
described, around 2024 the Hugging Face transformers library began
writing new models as edits of older ones, in modular files. This is
good engineering. No one should rewrite an attention block just to give
a new model its own file, and reuse is how cumulative work proceeds. It
also means that part of the rise in measured reuse after 2023 reflects
how the code is written, not only how models are designed. Our measure
counts an architecture when its code takes at least two of the
attention, feed-forward and decoder-layer blocks from another model.
That shows an implementation relationship. It says nothing about what a
block contributes to performance. The question for a report is whether
it tells readers what was reused and keeps that apart from what it
claims as new.

We conjecture that incentives also play a part, though our coding does
not measure them. A named change gives a report something to announce. A
controlled test costs compute that could otherwise go into the released
model. A team that must choose between a better model and a cleaner
experiment has an obvious reason to choose the model. The OLMo 2 and
Mistral 7B reports show that teams can make the other choice, at least
in part. We record what the reports say and test, not why their authors
made those choices.

Others have traced how models relate to one another. Laufer and
colleagues map model lineage across the Hugging Face ecosystem from
metadata (Laufer et al., 2025), HuggingGraph treats the same ecosystem
as a supply chain (Rahman et al., 2025), and Wei and colleagues use
lineage to identify disruptive models (Wei et al., 2026). Our measure
differs in its source: it reads the code of particular blocks, not the
metadata attached to a model. None of these three lines of work examines
the novelty claims of release reports and the evidence offered for them.
The shared-implementation tests ask whether modifications help, and the
lineage studies ask what builds on what. Neither asks whether a release
report's stated contribution is supported by the evidence the report
gives. A lineage record tells us what a report builds on. A controlled
test tells us what the claimed change adds.

Reporting tools exist, but they aim at other questions. Model cards give
a format for reporting models, datasheets do the same for datasets, and
the NeurIPS reproducibility checklist covers machine learning papers
(Mitchell et al., 2019; Gebru et al., 2021; Pineau et al., 2021). They
made reporting an explicit part of research practice, which is to their
credit. None asks a model release to separate new from adopted parts and
test the new ones. A report can therefore fill in every item of such a
format and still leave its novelty claim impossible to assess.

Taken together, these points say where the weakness lies. In the reports
whose attribution we read, the sources of adopted parts were credited,
as Section 3 showed. A contribution needs more than a favourable
comparison with another complete model. Readers need to know what was
changed and whether the comparison bears on that change. The weakness
appears wherever a report credits a gain to one changed part and offers,
as evidence, only a comparison between finished models.

\subsection{5. What good release research looks
like}\label{what-good-release-research-looks-like}

A release report does two jobs. It tells people what a model does and
how to use it, and it often claims that some choice made the model
better. The first job needs a clear account of the released system. The
second needs evidence that separates the effect of the claimed change
from everything else that changed. A report can do the first job well
and leave the second undone.

The rules in Box 1 respond to the gaps that Section 3 found. Novelty
claims often came without a test that bears on them. Results rarely came
with more than one run. Comparison scores were often copied from other
reports, and full disclosure of how a model was trained was rare. We
group the rules by their relation to our evidence. The first five match
items our coding measures, so we can say how often reports already
follow them. The last two rest on the argument alone, and we have no
counts to support them. These are rules for reporting. A release that
breaks them can still be useful.

\begin{center}\noindent\fbox{\begin{minipage}{0.95\linewidth}
\noindent\textbf{Box 1. Rules for release reports}\par\smallskip
\noindent\textit{Measured by our coding:}\par
\begin{enumerate}\setlength{\itemsep}{1pt}
\item[1.] For each part presented as new, report a controlled comparison with and without it, or say that none was run.
\item[2.] Report more than one run, or say that a result comes from a single run.
\item[3.] Rerun comparison models under one evaluation setup, and say which scores were copied from other reports.
\item[4.] Make the model available for outside evaluation, and report any evaluation by a third party.
\item[5.] Report the training data mixture, the token count and the compute.
\end{enumerate}
\noindent\textit{Proposed, not measured:}\par
\begin{enumerate}\setlength{\itemsep}{1pt}
\item[6.] Separate the parts the release introduces from those it adopts, give the origin of each adopted part, and give the established name next to any new name.
\item[7.] Describe the difference from the closest public baseline, meaning the most similar openly available model, in terms of the changes in code.
\end{enumerate}
\end{minipage}}\end{center}

The rules are easiest to judge on real reports. We apply them to two of
the coded reports: OLMoE, which met the most measured items, and Mistral
7B. Every quote below was found by script on the stated page.

The OLMoE report makes four architectural or training claims, for
example that fine-grained routing with many small experts matters and
that shared experts do not help, and each comes with a controlled
comparison (Muennighoff et al., 2024). One reads ``We vary the number of
experts in tandem with the FFN dimension to ensure that active and total
parameters'' stay matched (p.~10). The report does not say how many runs
stand behind its results. It states ``We run all evaluations ourselves
with 5 few-shots'' (p.~7), so no scores are copied, and it reports no
third-party evaluation. It gives the data mixture in a table (p.~5), the
token count (``train for a total of 5.133T tokens'', p.~6) and the
compute (``256 H100 GPUs for approximately 10 days'', p.~45). It meets
Rules 1, 3 and 5 and misses Rules 2 and 4.

The Mistral 7B report presents grouped-query attention and
sliding-window attention as the main features of the model and cites
earlier work for both (p.~1) (Jiang et al., 2023). Under Rule 6 these
are adopted parts, so Rule 1 asks only that the report say whether their
effect in this model was tested, and it reports no with/without
comparison. It does not state how many runs it used. It writes ``We
compare Mistral 7B to Llama, and re-run all benchmarks with our own
evaluation pipeline for fair comparison'' (p.~3) and mentions that
``independent human evaluation was conducted'' (p.~4). The training
data, token count and compute are not given. It meets Rules 3 and 4.

The two reports fall short in different places. A reader of the first
knows how the model was trained but not how stable the results are. A
reader of the second knows how the comparisons were made but not what
the model was trained on. Box 1 asks for both.

Rule 1 matters most to our argument. A comparison between finished
models tells a reader how the systems performed under a stated
evaluation. It does not say which choice produced the difference. A
controlled comparison changes the claimed part and holds model size,
data, and tokens fixed, so that a difference in the result can be read
against that part. Lipton and Steinhardt describe the wider habit in
machine learning of reporting gains without identifying where they come
from (Lipton and Steinhardt, 2018). When no controlled test was run, the
rule asks the authors to say so. That statement costs no compute, but it
is not free. A claim marked as untested reads as weaker than one left to
speak for itself, and Section 4 gave reasons why a team might prefer the
second. We accept that cost. The weaker statement is the accurate one,
and a claim presented as stronger than its evidence leaves the reader
with a wrong belief about what was shown.

Rules 2 and 3 concern how much weight a difference can bear. A single
run says nothing about how much a result would vary, and at pre-training
scale a second run may be out of reach. Stating that a result comes from
one run then tells the reader how to read it. Copied scores are
sometimes the only option, for instance when a comparison model cannot
be run. Marking them as copied lets a reader tell a difference between
models from a difference between evaluation procedures.

Rule 4 asks for what a release team can control. It cannot make an
outside group evaluate its model before publication. It can release the
model, or give access to it, in a form that others can test, and it can
report any outside evaluation that exists. Rule 5 asks for three facts
that answer different questions: what the data were, how much of them
was processed, and what resources stand behind the result. Where one of
them cannot be disclosed, the report can name the omission. A named
omission tells the reader which part of a comparison they cannot check,
while silence leaves them guessing.

Rule 6 asks for a practice that the reports we read already follow.
Section 3.2 looked for places where a report gives a new name to a
technique it says is already known. Among the five candidate cases the
coders proposed, none hid the origin of the component (Almazrouei et
al., 2023). We therefore do not present the naming clause as a
correction of a habit we observed. A new name placed next to the
established one costs nothing and lets a reader find the earlier work.
The clause belongs with the broader task of separating what is
introduced from what is adopted. That task comes before Rule 1. A reader
cannot ask for a test of a new part until the report says which part is
new.

Rule 7 gives that separation a concrete form. Section 3.1 compared five
models with Llama in the reference code, and in each case the difference
could be stated in a sentence or two (Touvron et al., 2023; Wolf et al.,
2020). A report can do the same against the most similar model that is
openly available. Such a description does not show that the changes
matter. It shows the reader where a novelty claim applies and which
comparison would test it.

Some reports already do much of this. The OLMo 2 report cites the origin
of each architectural change and shows its effect in a figure (OLMo et
al., 2025). The Mistral 7B report reran all comparison models with its
own evaluation pipeline (Jiang et al., 2023). On the five measured
items, the X-MOD, OLMoE and ELECTRA reports each met three, the most any
report met, while 24 of the 40 reports met none (Table A2 gives the
coding of each report). The score counts only what we could code, and it
is not a grade of the reports' quality or usefulness. It is also
stricter than Rule 1 in one respect. A report met the first item only if
every architectural or training claim it made had a controlled
comparison, and a statement that none was run did not count.

The rules leave room for releases that are not experiments. Combining
known techniques, training a model on them and making it available is
useful work. Authors who do this can say what they built, credit what
they used and report what they found. Readers can then judge the release
on those terms, without assuming that each choice was tested on its own.
We have no evidence from this study that the rules would improve
reports. We offer them because they make the grounds and the limits of a
claim visible.

Existing tools cover parts of this ground. Model cards document a model
for the people who use and study it (Mitchell et al., 2019), and
datasheets do the same for datasets (Gebru et al., 2021). The NeurIPS
reproducibility checklist sets out what a reader needs to assess and
repeat a piece of machine learning research (Pineau et al., 2021). Rules
2, 3, and 5 overlap with these tools, and a team that already fills them
in has done part of the work. None of them asks a release to name its
new parts and link each one to a test. That link is what Box 1 adds.

The audiences differ. Someone deciding whether to use a model needs to
know its intended use and its limits. Someone weighing a claimed
improvement needs to know what was changed and how the change was
tested. One report can serve both readers, provided it keeps apart the
sentences that describe the system and the sentences that state what an
experiment showed.

\subsection{6. Conclusions}\label{conclusions}

We asked a plain question of the release reports of large language
models: what does a report claim as new, and what evidence does it offer
for that claim? Section 3.3 gives the answer in full. Most of the claims
we coded come without a controlled comparison that would test them as
stated.

We saw no case in which a known component was renamed to hide its
origin, and the vendor reports we quoted credit their sources. The
difficulty lies elsewhere. Many releases change the architecture, the
data, the token budget and the training procedure at once, and then
compare the finished model with other finished models. Of the claims in
Section 3.3, 20 rest on a comparison that changes other things at the
same time and 34 rest on no comparison at all. A comparison of finished
models shows which one scored higher under the stated conditions. It
cannot say which change earned the credit.

Not every release works this way. The reports for X-MOD, OLMoE and
ELECTRA met more of our scored items than any others, which shows that a
release report can meet several of these standards at once. Section 5
describes two further reports whose practices a reader can check.

The advances traced in Section 2 set a fair standard. The transformer,
scaling laws, instruction tuning with human feedback, mixture of experts
and test-time compute were each presented together with a test of the
method (Vaswani et al., 2017; Kaplan et al., 2020; Ouyang et al., 2022;
Shazeer et al., 2017; Snell et al., 2024). A release report that
presents a component as new is making the same kind of claim. It should
come with the same kind of evidence.

The direct test of a claimed change is a run with and without it, with
size, data, and tokens held fixed. Where that run is too costly, or was
simply not done, the report can say so. Readers then know that the
result describes a system and not a component. A result from one run can
be labelled as such. A report that cannot give its data mixture, token
count or compute can say which of them it withholds.

Comparisons with other models remain worth making. They are worth more
when the releasing team reruns those models under one setup, and more
again when a third party evaluates the release. These practices tell a
user where a model stands, not why it stands there. A good report keeps
apart what the system does and what its experiments establish about its
parts.

Our evidence has limits. It comes from one library at one snapshot,
taken on 13 November 2026. The report sample could only include the 91
architectures whose documentation links an arXiv paper. The 80 without
such a link, which are the least documented, are outside it. Language
models coded the 40 reports, and a script checked every quote against
the source text, but no human has yet checked the coding. Agreement
between the model coders on the ablation level of matched claims was
moderate (Section 3.3), and on the data-mixture field it was low (kappa
0.30). The five code comparisons are illustrations, not a sample. We do
not measure model quality, and a claim that a report leaves untested may
still be true.

None of this asks release teams to stop building on each other's work. A
well-documented model trained from known parts is a useful thing to
publish. What we ask is narrower. When a report says that a part is new
and that it helps, it should include the run without that part, or one
sentence stating that this run was not done. A reader of a release
report should be able to see what is new, whether it was tested, and
what the model was trained on. In our sample, 19 of 73 novelty claims
came with a controlled test, and 1 of 40 reports gave its data mixture,
token count and compute together.

\subsection{Appendix A. Data and
method}\label{appendix-a.-data-and-method}

\subsubsection{A.1 Sampling frame and
sample}\label{a.1-sampling-frame-and-sample}

Frame: the 171 causal language-model architectures in the Hugging Face
transformers library at commit 0d7a621 (13 November 2026). For each, the
first arXiv identifier linked in its documentation page names its
report. 80 architectures link no arXiv paper; the remaining 91 link 82
distinct papers. Eligible: the paper names the model in its title or
abstract, dates from 2020 or later, and is not a paper about a single
component. 64 papers were eligible. We drew 40 at random (seed
20261114), stratified by the year of the arXiv identifier: 2020-2022: 9
of 15, 2023: 5 of 8, 2024: 14 of 23, 2025: 10 of 16, 2026: 2 of 2. All
40 full texts were taken from arXiv. Table S1 lists the whole frame with
each eligibility decision.

\subsubsection{A.2 Coding}\label{a.2-coding}

The coding protocol was written before any report was read. Two coders,
Claude Opus 5.5 and OpenAI GPT-6.1 Sol, coded each report independently
from the same page-numbered text and instructions, giving a verbatim
quote and page for every value. A script checked every quote against the
PDF text: a value whose quote was not found fell to the lowest level,
and a claim whose quote was not found was dropped. The two coders' claim
lists were matched by wording and page (51 matched claims); the 53
claims listed by one coder only, and every disagreement, went to the
third model. Disagreements went to xAI Grok 4.7, chosen because it
belongs to neither coder's family and its developer has no report in the
sample; its quotes were checked in the same way. No human check of the
coding has been done yet. Table S2 lists every coded claim with its
verbatim quote and page.

Table A1. Agreement between the two model coders, by coded field
(Cohen's kappa, raw agreement and number of items).

{\def\LTcaptype{none} 
\begin{longtable}[]{@{}llll@{}}
\toprule\noalign{}
Field & Cohen's kappa & Raw agreement & n \\
\midrule\noalign{}
\endhead
\bottomrule\noalign{}
\endlastfoot
More than one run & 0.78 & 0.95 & 40 \\
Comparators re-run & 0.77 & 0.85 & 40 \\
Scores copied & 0.73 & 0.82 & 40 \\
Third-party evaluation & 0.75 & 0.93 & 40 \\
Data mixture & 0.30 & 0.75 & 40 \\
Token count & 0.81 & 0.93 & 40 \\
Compute & 0.88 & 0.95 & 40 \\
Ablation level (matched claims) & 0.67 & 0.78 & 51 \\
\end{longtable}
}

\subsubsection{A.3 Coded reports}\label{a.3-coded-reports}

\begin{landscape}\scriptsize

Table A2. The 40 coded reports: architectural and training novelty
claims by evidence level (controlled / confounded / none), runs, re-run
comparators, copied scores, third-party evaluation, disclosure, and the
number of the five measured items met.

{\def\LTcaptype{none} 
\begin{longtable}[]{@{}
  >{\raggedright\arraybackslash}p{(\linewidth - 18\tabcolsep) * \real{0.1000}}
  >{\raggedright\arraybackslash}p{(\linewidth - 18\tabcolsep) * \real{0.1000}}
  >{\raggedright\arraybackslash}p{(\linewidth - 18\tabcolsep) * \real{0.1000}}
  >{\raggedright\arraybackslash}p{(\linewidth - 18\tabcolsep) * \real{0.1000}}
  >{\raggedright\arraybackslash}p{(\linewidth - 18\tabcolsep) * \real{0.1000}}
  >{\raggedright\arraybackslash}p{(\linewidth - 18\tabcolsep) * \real{0.1000}}
  >{\raggedright\arraybackslash}p{(\linewidth - 18\tabcolsep) * \real{0.1000}}
  >{\raggedright\arraybackslash}p{(\linewidth - 18\tabcolsep) * \real{0.1000}}
  >{\raggedright\arraybackslash}p{(\linewidth - 18\tabcolsep) * \real{0.1000}}
  >{\raggedright\arraybackslash}p{(\linewidth - 18\tabcolsep) * \real{0.1000}}@{}}
\toprule\noalign{}
\begin{minipage}[b]{\linewidth}\raggedright
Model (library name)
\end{minipage} & \begin{minipage}[b]{\linewidth}\raggedright
arXiv
\end{minipage} & \begin{minipage}[b]{\linewidth}\raggedright
Stratum
\end{minipage} & \begin{minipage}[b]{\linewidth}\raggedright
ARCH/TRAIN claims (controlled / confounded / none)
\end{minipage} & \begin{minipage}[b]{\linewidth}\raggedright
Runs
\end{minipage} & \begin{minipage}[b]{\linewidth}\raggedright
Re-run
\end{minipage} & \begin{minipage}[b]{\linewidth}\raggedright
Copied
\end{minipage} & \begin{minipage}[b]{\linewidth}\raggedright
Third party
\end{minipage} & \begin{minipage}[b]{\linewidth}\raggedright
Data / tokens / compute
\end{minipage} & \begin{minipage}[b]{\linewidth}\raggedright
Items met
\end{minipage} \\
\midrule\noalign{}
\endhead
\bottomrule\noalign{}
\endlastfoot
big\_bird & 2007.14062 & 2020-2022 & 2 (0/1/1) & VAR & MIXED & MIXED &
YES & PARTIAL / N / N & 2 of 5 \\
codegen & 2203.13474 & 2020-2022 & 0 (0/0/0) & NR & MIXED & MIXED & NO &
PARTIAL / Y / N & 0 of 4 \\
electra & 2003.10555 & 2020-2022 & 2 (2/0/0) & MULTI & MIXED & MIXED &
YES & PARTIAL / Y / Y & 3 of 5 \\
gpt\_neox & 2204.06745 & 2020-2022 & 0 (0/0/0) & NR & MIXED & MIXED & NO
& NONE / N / Y & 0 of 4 \\
mbart & 2001.08210 & 2020-2022 & 1 (0/1/0) & NR & NO & MIXED & NO & NONE
/ N / Y & 0 of 5 \\
mvp & 2206.12131 & 2020-2022 & 2 (0/0/2) & NR & NR & MIXED & NO & FULL /
N / N & 0 of 5 \\
opt & 2205.01068 & 2020-2022 & 0 (0/0/0) & NR & MIXED & YES & NO &
PARTIAL / Y / N & 0 of 4 \\
prophetnet & 2001.04063 & 2020-2022 & 2 (0/1/1) & NR & NR & NR & NO &
PARTIAL / N / N & 0 of 5 \\
xmod & 2205.06266 & 2020-2022 & 1 (1/0/0) & MULTI & YES & NR & NO &
PARTIAL / N / N & 3 of 5 \\
falcon & 2311.16867 & 2023 & 0 (0/0/0) & VAR & NO & YES & YES & PARTIAL
/ Y / Y & 2 of 4 \\
kosmos2\_5 & 2309.11419 & 2023 & 1 (0/0/1) & NR & NR & NR & NO & PARTIAL
/ N / N & 0 of 5 \\
mamba & 2312.00752 & 2023 & 2 (2/0/0) & NR & NR & NR & NO & PARTIAL / Y
/ N & 1 of 5 \\
mistral & 2310.06825 & 2023 & 2 (0/0/2) & NR & YES & NO & YES & NONE / N
/ N & 2 of 5 \\
phi & 2306.11644 & 2023 & 0 (0/0/0) & NR & MIXED & YES & NO & PARTIAL /
Y / Y & 0 of 4 \\
aria & 2410.05993 & 2024 & 3 (0/0/3) & NR & MIXED & MIXED & NO & PARTIAL
/ Y / N & 0 of 5 \\
deepseek\_v3 & 2412.19437 & 2024 & 3 (2/1/0) & NR & YES & NO & NO &
PARTIAL / Y / Y & 1 of 5 \\
emu3 & 2409.18869 & 2024 & 1 (0/0/1) & NR & MIXED & MIXED & NO & PARTIAL
/ N / N & 0 of 5 \\
falcon\_mamba & 2410.05355 & 2024 & 2 (0/1/1) & NR & MIXED & MIXED & YES
& PARTIAL / Y / N & 1 of 5 \\
gemma & 2403.08295 & 2024 & 0 (0/0/0) & NR & MIXED & MIXED & NO & NONE /
N / N & 0 of 4 \\
gemma2 & 2408.00118 & 2024 & 2 (1/0/1) & NR & MIXED & MIXED & YES &
PARTIAL / Y / N & 1 of 5 \\
mamba2 & 2405.21060 & 2024 & 4 (1/2/1) & NR & NR & NR & NO & PARTIAL / Y
/ N & 0 of 5 \\
moshi & 2410.00037 & 2024 & 4 (3/0/1) & NR & MIXED & MIXED & NO &
PARTIAL / Y / N & 0 of 5 \\
nemotron & 2402.16819 & 2024 & 0 (0/0/0) & NR & MIXED & YES & NO &
PARTIAL / Y / Y & 0 of 4 \\
olmo & 2402.00838 & 2024 & 0 (0/0/0) & NR & NR & NR & NO & NONE / Y / N
& 0 of 4 \\
olmoe & 2409.02060 & 2024 & 4 (4/0/0) & NR & YES & NO & NO & FULL / Y /
Y & 3 of 5 \\
phi3 & 2404.14219 & 2024 & 0 (0/0/0) & NR & MIXED & MIXED & NO & PARTIAL
/ Y / N & 0 of 4 \\
recurrent\_gemma & 2404.07839 & 2024 & 1 (0/1/0) & NR & NR & NR & NO &
NONE / Y / N & 0 of 5 \\
zamba2 & 2411.15242 & 2024 & 4 (0/0/4) & NR & NR & NR & NO & PARTIAL / Y
/ Y & 0 of 5 \\
bitnet & 2504.12285 & 2025 & 0 (0/0/0) & NR & YES & NO & NO & PARTIAL /
Y / N & 1 of 4 \\
dots1 & 2506.05767 & 2025 & 0 (0/0/0) & NR & YES & NO & NO & PARTIAL / Y
/ Y & 1 of 4 \\
exaone4 & 2507.11407 & 2025 & 1 (0/0/1) & NR & MIXED & MIXED & NO & NONE
/ Y / Y & 0 of 5 \\
glm4\_moe & 2508.06471 & 2025 & 2 (0/0/2) & NR & NR & NR & NO & PARTIAL
/ Y / N & 0 of 5 \\
hrm\_text & 2506.21734 & 2025 & 5 (0/3/2) & NR & MIXED & MIXED & YES &
PARTIAL / N / N & 1 of 5 \\
kimi\_linear & 2510.26692 & 2025 & 2 (1/1/0) & NR & YES & NO & NO &
PARTIAL / Y / N & 1 of 5 \\
longcat\_flash & 2509.01322 & 2025 & 9 (0/6/3) & NR & MIXED & MIXED & NO
& PARTIAL / Y / Y & 0 of 5 \\
minimax & 2501.08313 & 2025 & 3 (1/1/1) & NR & MIXED & MIXED & NO &
PARTIAL / Y / N & 0 of 5 \\
phi4\_multimodal & 2503.01743 & 2025 & 2 (0/0/2) & NR & MIXED & MIXED &
YES & PARTIAL / Y / N & 1 of 5 \\
youtu & 2512.24618 & 2025 & 3 (1/1/1) & NR & YES & NO & NO & PARTIAL / Y
/ N & 1 of 5 \\
deepseek\_v4 & 2606.19348 & 2026 & 3 (0/0/3) & NR & MIXED & NR & NO &
PARTIAL / Y / N & 0 of 5 \\
solar\_open & 2601.07022 & 2026 & 0 (0/0/0) & NR & NR & NR & NO &
PARTIAL / Y / N & 0 of 4 \\
\end{longtable}
}

\end{landscape}

All scripts, codings, quotes and decisions are in Supplementary Material
S1.

\subsection{Declarations}\label{declarations}

Data and code availability. The sampling frame, eligibility decisions,
sample, both coders' outputs with verbatim quotes and page numbers, the
resolution outputs, the agreement analysis, the code comparisons and all
scripts are provided as Supplementary Material S1 (archive with a
README), including Table S1 (the sampling frame with every eligibility
decision) and Table S2 (all 73 coded claims with verbatim quotes and
pages). The report PDFs are not redistributed; each is identified by its
arXiv identifier.

Use of AI tools. Language models were used in two ways. As research
instruments, two models coded the release reports and a third settled
disagreements, as described in Section 3.3 and Appendix A; every coded
value was checked by script against the source text. As writing aids,
language models (Claude, Anthropic; GPT models, OpenAI) were used to
draft and edit text from the author's claim ledger and outline. The
author checked every number, quote and reference against the data and
the sources, and takes full responsibility for the content.

Competing interests. The author declares no competing interests.

Funding. No specific funding was received for this work.

\subsection{References}\label{references}

Ainslie, Lee-Thorp, de Jong, Zemlyanskiy, Lebrón, Sanghai (2023). GQA:
Training Generalized Multi-Query Transformer Models from Multi-Head
Checkpoints. In: Proceedings of the 2023 Conference on Empirical Methods
in Natural Language Processing, 4895-4901. arXiv:2305.13245.
https://doi.org/10.18653/v1/2023.emnlp-main.298

Almazrouei, Alobeidli, Alshamsi, Cappelli, Cojocaru, Debbah et
al.~(2023). The Falcon Series of Open Language Models. arXiv:2311.16867.

Beltagy, Peters, Cohan (2020). Longformer: The Long-Document
Transformer. arXiv:2004.05150.

Bouthillier, Delaunay, Bronzi, Trofimov, Nichyporuk, Szeto et
al.~(2021). Accounting for Variance in Machine Learning Benchmarks.
arXiv:2103.03098.

Brown, Mann, Ryder, Subbiah, Kaplan, Dhariwal et al.~(2020). Language
Models are Few-Shot Learners. arXiv:2005.14165.

Camacho-Villalón, Dorigo, Stützle (2023). Exposing the grey wolf,
moth-flame, whale, firefly, bat, and antlion algorithms: six misleading
optimization techniques inspired by bestial metaphors. International
Transactions in Operational Research 30(6), 2945-2971.
https://doi.org/10.1111/itor.13176

Child, Gray, Radford, Sutskever (2019). Generating Long Sequences with
Sparse Transformers. arXiv:1904.10509.

Dao, Gu (2024). Transformers are SSMs: Generalized Models and Efficient
Algorithms Through Structured State Space Duality. arXiv:2405.21060.

Dehghani, Djolonga, Mustafa, Padlewski, Heek, Gilmer et al.~(2023).
Scaling Vision Transformers to 22 Billion Parameters. arXiv:2302.05442.

Devlin, Chang, Lee, Toutanova (2019). BERT: Pre-training of Deep
Bidirectional Transformers for Language Understanding. In: Proceedings
of the 2019 Conference of the North American Chapter of the Association
for Computational Linguistics: Human Language Technologies, Volume 1
(Long and Short Papers), 4171-4186. arXiv:1810.04805.
https://doi.org/10.18653/v1/n19-1423

Dodge, Gururangan, Card, Schwartz, Smith (2019). Show Your Work:
Improved Reporting of Experimental Results. In: Proceedings of the 2019
Conference on Empirical Methods in Natural Language Processing and the
9th International Joint Conference on Natural Language Processing
(EMNLP-IJCNLP), 2185-2194. arXiv:1909.03004.
https://doi.org/10.18653/v1/d19-1224

Fedus, Zoph, Shazeer (2022). Switch Transformers: Scaling to Trillion
Parameter Models with Simple and Efficient Sparsity. arXiv:2101.03961.

Gebru, Morgenstern, Vecchione, Vaughan, Wallach, Daumé et al.~(2021).
Datasheets for Datasets. arXiv:1803.09010.

Gemma Team, Mesnard, Hardin, Dadashi, Bhupatiraju, Pathak et al.~(2024).
Gemma: Open Models Based on Gemini Research and Technology.
arXiv:2403.08295.

Gu, Dao (2023). Mamba: Linear-Time Sequence Modeling with Selective
State Spaces. arXiv:2312.00752.

Hoffmann, Borgeaud, Mensch, Buchatskaya, Cai, Rutherford et al.~(2022).
Training Compute-Optimal Large Language Models. arXiv:2203.15556.

Jiang, Sablayrolles, Mensch, Bamford, Chaplot, Casas et al.~(2023).
Mistral 7B. arXiv:2310.06825.

Kaplan, McCandlish, Henighan, Brown, Chess, Child et al.~(2020). Scaling
Laws for Neural Language Models. arXiv:2001.08361.

Laufer, Oderinwale, Kleinberg (2025). Anatomy of a Machine Learning
Ecosystem: 2 Million Models on Hugging Face. arXiv:2508.06811.

Lipton, Steinhardt (2018). Troubling Trends in Machine Learning
Scholarship. arXiv:1807.03341.

Melis, Dyer, Blunsom (2018). On the State of the Art of Evaluation in
Neural Language Models. arXiv:1707.05589.

Mitchell, Wu, Zaldivar, Barnes, Vasserman, Hutchinson et al.~(2019).
Model Cards for Model Reporting. In: Proceedings of the Conference on
Fairness, Accountability, and Transparency, 220-229. arXiv:1810.03993.
https://doi.org/10.1145/3287560.3287596

Muennighoff, Soldaini, Groeneveld, Lo, Morrison, Min et al.~(2024).
OLMoE: Open Mixture-of-Experts Language Models. arXiv:2409.02060.

Narang, Chung, Tay, Fedus, Fevry, Matena et al.~(2021). Do Transformer
Modifications Transfer Across Implementations and Applications? In:
Proceedings of the 2021 Conference on Empirical Methods in Natural
Language Processing, 5758-5773. arXiv:2102.11972.
https://doi.org/10.18653/v1/2021.emnlp-main.465

OLMo, Walsh, Soldaini, Groeneveld, Lo, Arora et al.~(2025). 2 OLMo 2
Furious. arXiv:2501.00656.

Ouyang, Wu, Jiang, Almeida, Wainwright, Mishkin et al.~(2022). Training
language models to follow instructions with human feedback.
arXiv:2203.02155.

Pineau, Vincent-Lamarre, Sinha, Larivière, Beygelzimer, d'Alché-Buc et
al.~(2021). Improving Reproducibility in Machine Learning Research (A
Report from the NeurIPS 2019 Reproducibility Program). arXiv:2003.12206.

Rahman, Gao, Ji (2025). HuggingGraph: Understanding the Supply Chain of
LLM Ecosystem. In: Proceedings of the 34th ACM International Conference
on Information and Knowledge Management, 5997-6005. arXiv:2507.14240.
https://doi.org/10.1145/3746252.3761510

Shazeer, Mirhoseini, Maziarz, Davis, Le, Hinton et al.~(2017).
Outrageously Large Neural Networks: The Sparsely-Gated
Mixture-of-Experts Layer. arXiv:1701.06538.

Shazeer (2019). Fast Transformer Decoding: One Write-Head is All You
Need. arXiv:1911.02150.

Shazeer (2020). GLU Variants Improve Transformer. arXiv:2002.05202.

Snell, Lee, Xu, Kumar (2024). Scaling LLM Test-Time Compute Optimally
can be More Effective than Scaling Model Parameters. arXiv:2408.03314.

Sörensen (2015). Metaheuristics---the metaphor exposed. International
Transactions in Operational Research 22(1), 3-18.
https://doi.org/10.1111/itor.12001

Touvron, Lavril, Izacard, Martinet, Lachaux, Lacroix et al.~(2023).
LLaMA: Open and Efficient Foundation Language Models. arXiv:2302.13971.

Vaswani, Shazeer, Parmar, Uszkoreit, Jones, Gomez et al.~(2017).
Attention Is All You Need. arXiv:1706.03762.

Wei, Kang, Pan, Yang (2026). Identifying Disruptive Models in the
Open-Source LLM Community. arXiv:2604.11618.

Weyland (2012). A Rigorous Analysis of the Harmony Search Algorithm. In:
Modeling, Analysis, and Applications in Metaheuristic Computing, 72-83.
https://doi.org/10.4018/978-1-4666-0270-0.ch005

Wolf, Debut, Sanh, Chaumond, Delangue, Moi et al.~(2020). HuggingFace's
Transformers: State-of-the-art Natural Language Processing.
arXiv:1910.03771.

Yang, Yang, Hui, Zheng, Yu, Zhou et al.~(2024). Qwen2 Technical Report.
arXiv:2407.10671.

Yang, Li, Yang, Zhang, Hui, Zheng et al.~(2025). Qwen3 Technical Report.
arXiv:2505.09388.

Zhao, Lu, Huang, Zhang, Zhou (2026). Most Transformer Modifications
Still Do Not Transfer at 1-3B: A 2020-2026 Update to Narang et
al.~(2021) with Downstream Evaluation and a Noise Floor.
arXiv:2605.20798.

Zuo, Velikanov, Rhaiem, Chahed, Belkada, Kunsch et al.~(2024). Falcon
Mamba: The First Competitive Attention-free 7B Language Model.
arXiv:2410.05355.

\end{document}